\documentclass{article} 
\usepackage[preprint]{neurips_2026}

\usepackage{amsmath,amsfonts,bm}

\def\eqref#1{equation~\ref{#1}}

\def\1{\bm{1}}

\DeclareMathAlphabet{\mathsfit}{\encodingdefault}{\sfdefault}{m}{sl}
\SetMathAlphabet{\mathsfit}{bold}{\encodingdefault}{\sfdefault}{bx}{n}

\usepackage{hyperref}
\usepackage{url}
\usepackage{array}
\usepackage{booktabs}
\usepackage{graphicx}
\usepackage{float}
\usepackage{multirow}
\usepackage{xcolor}
\usepackage{framed}
\definecolor{shadecolor}{gray}{0.95}

\newcommand{\methodname}{RoleWorld} 

\title{RoleWorld: Knowledge Transfer in World Modeling with Functional Role Abstractions}

\author{%
  Boyuan Xiao \\
  {\normalfont State Key Lab of CAD\&CG} \\
  {\normalfont Zhejiang University, China} \\
  \And
  Pu Wang \\
  {\normalfont State Key Lab of CAD\&CG} \\
  {\normalfont Zhejiang University, China} \\
  \AND
  Ji Feng \\
  {\normalfont Baiont Quant, } \\
  {\normalfont China} \\
  \And
  Yao-Xiang Ding \\
  {\normalfont State Key Lab of CAD\&CG} \\
  {\normalfont Zhejiang University, China} \\
}

\begin{document}

\maketitle

\begin{abstract}
World models enable large language model (LLM) agents to anticipate action effects and make informed decisions. For generalization, it is desirable for such world models to transfer efficiently to new environments with similar interaction structure and underlying dynamics. Recent approaches encode world dynamics as symbolic rules or executable programs, whose compositional structure in principle affords transfer, but still organize such knowledge around environment-specific objects. To address this limitation, we introduce \emph{\methodname{}}, a symbolic world modeling approach which uses reusable prototypes and environment-specific parameter values as \emph{functional role abstractions} to represent how objects behave in interactions. From interaction trajectories, a teacher LLM extracts role prototypes and executable transition rules defined over role instances, together with a skill that guides the actor LLM to use role information in decision making. 
In a target world, without the teacher, the actor uses this skill together with its current role assignments and parameter estimates, and actively tests a role instance hypothesis through interaction when its uncertainty affects the current decision. Feedback from each action refines role assignments and estimates parameter values online, with no LLM parameter updates and the transition rules held fixed. Experiments on MARS benchmark show that \methodname{} enables robust generalization across environments with significantly altered object behavior, suggesting that functional role abstractions can serve as an effective component for knowledge transfer in world modeling. 
Project Page: https://future-item.github.io/RoleWorld.

\end{abstract}

\section{Introduction}

Learning how the world works enables an agent to anticipate the consequences of its actions and make informed decisions \citep{schrittwieser2020mastering}. World models capture such knowledge by describing how actions change the environment \citep{ha2018world,hafner2025mastering}, allowing agents, such as Large Language Models (LLMs), to reason about possible outcomes before acting \citep{zhou2026wall}. Recent evaluations increasingly consider agents across multiple related environments or environment variants, rather than within a single fixed world \citep{tang2024mars,tang2024worldcoder}. This setting creates a natural need for world models that can reuse previously acquired knowledge when entering a new environment, instead of learning its dynamics entirely from scratch \citep{taylor2009transfer}. Such transfer is plausible because related environments often preserve substantial interaction structure and underlying dynamics even as their particular configurations vary \citep{tang2024worldcoder}. A central question is therefore \emph{how to represent world knowledge} so that it can transfer efficiently across related environments.

Recent approaches use LLMs to express world dynamics as symbolic rules or executable programs \citep{zhou2026wall,tang2024worldcoder,piriyakulkij2026poe}, whose compositional structure provides a natural basis for reuse \citep{battaglia2018relational}. 
However, their rules remain tied to environment-specific objects and can fail when familiar objects serve different functions. This suggests a functional view, as in affordance models relating objects, actions, and effects \citep{montesano2008learning,khetarpal2026affordances}. We organize such functional knowledge for transfer by separating shared interaction mechanisms from the object--function assignments and parameters that can be revised online.

To address this problem, we introduce \emph{\methodname{}}, a symbolic world modeling approach that uses \emph{functional role abstractions} to represent how objects behave in interactions. Each functional role abstraction consists of a reusable \emph{role prototype} and environment-specific \emph{parameter values}. This factorization separates reusable interaction functions from environment-specific object–function assignments and parameter values. As a result, the same role-level interaction structure can be reused when the objects realizing a function change across environments, while environment-specific differences are captured through role assignments and parameters. From trajectories collected in a source world, a teacher LLM extracts role prototypes and executable \emph{transition rules} defined over role instances, together with a reusable \emph{skill} for role-based decision making. Each transition rule captures an interaction mechanism whose requirements and effects are instantiated through concrete role instances. This formulation is intended to support reuse across environments where different objects instantiate the same role prototypes. 
In a target world, the actor LLM operates without the teacher, and the role prototypes, transition rules, and skill remain fixed. Guided by the skill, the actor uses the transition rules to predict action outcomes and select actions. When uncertainty about a role assignment could affect its decision, the actor gathers evidence through an informative interaction. Action feedback is then used to refine role assignments and parameter estimates online, allowing \methodname{} to adapt the environment-specific instantiation of its world model.

We evaluate \methodname{} on MARS \citep{tang2024mars}, a challenging benchmark for evaluating generalization in world modeling. The results show that \methodname{} enables robust generalization across environments with significantly altered object behavior, suggesting that functional role abstractions can serve as an effective component for knowledge transfer in world modeling.

\section{Related Work}

LLM world models support look-ahead planning through prompted predictions and in-context dynamics estimation \citep{hao2023reasoning,liu2024reason}. Induction from Reflection (IFR) and WALL-E induce environment rules from interaction, with WALL-E learning executable rules to correct predictions \citep{tang2024mars,zhou2026wall}. PoE-World composes programmatic experts organized by object type, while OneLife models stochastic dynamics through conditionally activated laws \citep{piriyakulkij2026poe,khan2026one}.

Object-centric abstractions support reuse beyond individual object identities. OO-MDPs and Schema Networks represent interaction rules and local causal schemas \citep{diuk2008object,kansky2017schema}. \citet{james2022autonomous} learn object types and lifted planning models reusable across tasks sharing those types, while \citet{marom2024transferable} learn transferable transition and reward models. Adaptation can involve revising transferred code, as in WorldCoder \citep{tang2024worldcoder}, or inferring task-specific hidden parameters of shared dynamics, as in HiP-MDPs \citep{doshi2016hidden}. \methodname{} allows multiple compatible roles per object and adapts their assignments and instance parameters while holding role prototypes and transition rules fixed. We consider target-world changes expressible within this transferred library.

Affordance models provide a functional view of objects through object--action--effect relations \citep{montesano2008learning} and support partial transition modeling \citep{khetarpal2020can}. \citet{khetarpal2026affordances} extend affordances to multiple tasks and improve search with LLM partial models. SR-WM \citep{cheng2026beyond} binds visual entities to task-specific roles such as gripper and target. Our roles describe reusable interaction functions with parameterized requirements and effects. Transferring these functions still requires inferring which objects instantiate them and with what parameter values in the new environment.

Active model learning offers a way to resolve such uncertainty by selecting interactions that test hypotheses. WorldLLM \citep{levy2025worldllm} combines Bayesian inference over natural-language hypotheses with curiosity-driven exploration. OPINE-World \citep{courtis2026opine} combines program synthesis, hypothesis testing, and replay verification, using a Bayesian measure of object-type adequacy to guide exploration. In \methodname{}, a fixed skill guides such testing when uncertainty about a role instance hypothesis affects the current decision. Feedback refines role assignments and instance parameters while the prototypes and transition rules remain fixed, without further teacher calls.

\section{Method}

\methodname{} uses functional role abstractions to separate reusable interaction structure from environment-specific instantiations. Our framework consists of three components. First, a teacher LLM constructs a symbolic world model in the source world, consisting of reusable role prototypes and executable transition rules defined over role instances. Second, in a target world, we infer and progressively refine object--role assignments and environment-specific parameter values from interaction feedback, while keeping the prototypes and transition rules fixed. Third, a reusable skill guides the actor to make decisions using current role estimates and transition rules, and to take actions to test those estimates when role uncertainty affects the current decision.

\begin{figure}[!t]
    \centering
    \includegraphics[width=0.9\linewidth]{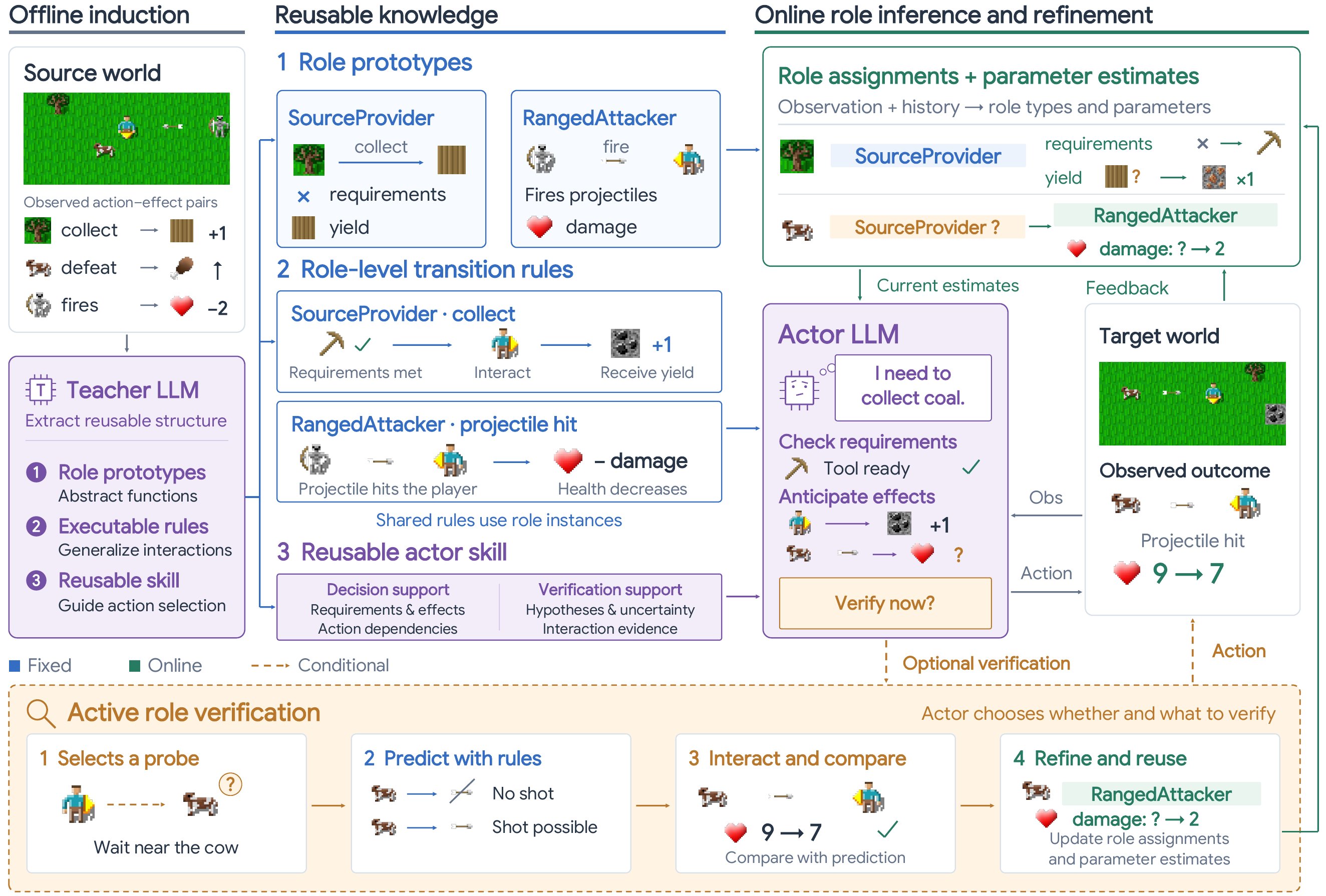}
    \caption{\textbf{Overview of \methodname{}}. A teacher LLM induces role prototypes, transition rules, and an actor skill from source trajectories. These remain fixed in the target world, where the actor refines role estimates online and tests hypotheses when uncertainty could affect its decision.}
    \label{fig:pipeline}
\end{figure}

\subsection{Offline World Model Construction}

Given source-world trajectories, we use an LLM teacher to jointly construct a role-based world model $\mathcal{M}=(\mathcal{R},\mathcal{T})$, where $\mathcal{R}$ is a library of reusable role prototypes and $\mathcal{T}$ is a set of executable transition rules over instances of these prototypes. The teacher constructs these rules from interaction evidence using predefined event types and operators in a domain-specific language (DSL).

The teacher proposes candidate functions and jointly specifies their role prototypes and transition rules, including triggering events, participants, conditions, effects, and the required parameters and state variables. It executes candidate rules to compare their predicted observable outcomes with real trajectories and obtains additional evidence through environment-interaction tools when needed. To resolve discrepancies, it revises parameter values or rule conditions and effects, adding or splitting prototypes and their associated rules when an independent function requires a separate representation. It merges candidate prototypes and their transition rules when they share triggering events, participant structure, and condition--effect structure, with differences in concrete requirements and effects captured by the parameter values of role instances. Each structural revision is checked against accumulated interaction evidence. Appendix~\ref{sec:offline_construction_details} describes the construction procedure.

\paragraph{Role prototypes.}
A role prototype is the reusable component of a functional role abstraction: it defines a behavioral function that different concrete objects can instantiate. We represent it as
\[
\rho =
\bigl(
\mathcal{C}_{\rho},
\Theta_{\rho},
\mathcal{X}_{\rho}
\bigr),
\]
where $\mathcal{C}_{\rho}$ is the set of role prototypes that can be instantiated alongside $\rho$ on the same object; $\Theta_{\rho}$ and $\mathcal{X}_{\rho}$ define the parameter and state spaces of its instances. An object may instantiate multiple compatible prototypes when it supports multiple functions.

\paragraph{Role instances.}
A role instance is the instantiation of a role prototype $\rho$ for a specific object in an environment $e$, with parameter values $\theta_{\rho}^{(e)} \in \Theta_{\rho}$ and, at each interaction step $t$, a state $x_{\rho,t}^{(e)} \in \mathcal{X}_{\rho}$, where the object index is omitted for brevity. A role assignment associates an object with a prototype; the parameter values specify the resulting instance's interaction requirements and effects. These values may differ across objects and environments but remain fixed for a given object within an environment. The values of the state variables may change at each interaction step. We constrain the teacher to declare only parameters and state variables that are referenced by at least one transition condition or effect.

\paragraph{Role-level transition rules.}
Transition rules describe interaction dynamics over role instances. The teacher constructs executable transition rules in the DSL, defining their conditions and effects in terms of the instances' parameters and states, as illustrated in Figure~\ref{fig:pipeline}. We represent a transition rule as
\[
\tau =
\bigl(
\mathrm{event},
P_{\tau},
g_{\tau},
F_{\tau}
\bigr),
\]
where \(\mathrm{event}\) is the triggering event, \(P_{\tau}\) specifies participant variables and their required role prototypes, \(g_{\tau}\) specifies the conditions under which the rule applies, and \(F_{\tau}\) specifies the resulting state changes and interaction outcomes. At execution time, participant variables bind role instances of the required prototypes. Parameters specialize a reusable rule for a particular role instance, while states track properties that evolve through interaction. For stochastic interactions, a rule specifies a distribution over trajectory-supported outcomes, with the distribution expressed in terms of the instances' parameters and states. Because transition rules reference role prototypes and the parameters and states of their instances rather than concrete object identities, the same rule can be instantiated by different objects with different parameter values across environments.

\subsection{Online Role Inference and Refinement}
With role prototypes and transition rules held fixed, online role inference estimates object--role assignments and environment-specific role parameters from interaction feedback. \methodname{} initializes these role estimates from the role assignments and parameter values retained in the source world, treating them as revisable priors. For each encountered object, it maintains role instance hypotheses with associated probabilities. Each role instance hypothesis is a candidate instantiation for an object, specifying a role type, identified by its prototype, and a particular set of parameter values. For each interaction, we evaluate combinations of role instance hypotheses, considering only the prototypes and parameters referenced by the relevant transition rules. We predict the interaction effects by directly executing these DSL rules with the current states and action under each evaluated combination. We express the effects recorded in environment feedback in the same DSL operation representation and compare them with the predictions to update hypothesis probabilities and guide revisions to role assignments and parameter values.

\paragraph{Distinguishing parameter errors from role assignment errors.}
A prediction error does not necessarily imply an incorrect role type, since interaction outcomes also depend on environment-specific parameters. If the predicted and observed DSL operations have the same categories and multiplicities but differ in their concrete argument values---for example, predicting a yield of wood while observing a yield of coal (see Figure~\ref{fig:pipeline})---the mismatch can still be explained by incorrect parameter values. Here, an operation category identifies the operation kind and, for state changes or resets, the affected field. The model therefore first revises the parameters referenced by the responsible rules, considering both existing alternatives and new values suggested by feedback. Only when parameter revision cannot explain a category-level discrepancy does the model reconsider the role assignments. It then considers additions, removals, or substitutions of assigned prototypes within the existing library, subject to compatibility constraints, and jointly re-evaluates the parameter values in the resulting role instance hypotheses.

\paragraph{Updating role instance hypothesis probabilities from interaction feedback.}
After identifying whether a prediction discrepancy calls for a parameter or role assignment revision, \methodname{} evaluates the resulting combinations of role instance hypotheses against the observed interaction. Let $H$ denote a combination of compatible role instance hypotheses for the objects participating in the interaction. We compare each predicted outcome $o$ with the observed outcome $o_t^{\mathrm{obs}}$ as multisets of DSL operations. Before comparison, repeated state and quantity changes are consolidated into their net effects. An operation matches only when its kind and all argument values agree. Let $\mathcal{O}(H)$ denote the possible outcomes predicted under $H$; deterministic rules predict a single outcome with probability one. We define the feedback score as
\[
\mathcal{L}_t(H)
=
\sum_{o \in \mathcal{O}(H)}
P(o \mid H)\,
\mathbf{1}[o=o_t^{\mathrm{obs}}],
\]
where the indicator requires an exact match of the operation multisets.
The resulting feedback score reweights the probability of each combination as
\begin{equation}
    P_t(H) \propto P_{t-1}(H)\,\mathcal{L}_t(H),
\end{equation}
with normalization over the evaluated combinations. If no evaluated hypothesis combination assigns positive probability to the observed outcome, we treat the observation as unexplained and trigger hypothesis expansion or revision rather than performing the probability update.

The initial probabilities of hypothesis combinations are approximated from those of their constituent role instance hypotheses. To bound inference, each update evaluates only a finite set of high-probability combinations rather than the full Cartesian product of hypothesis sets. Marginal confidence in an individual role instance hypothesis is obtained by summing the updated probabilities of the combinations that contain it.

\paragraph{Validating revisions against accumulated evidence.}
After updating hypothesis probabilities, we determine whether to accept a proposed revision to role assignments or parameter values by checking it against the complete interaction history. A revision from the current hypothesis combination $H$ to a proposed combination $H'$ is accepted only if it improves the cumulative log feedback score:
\begin{equation}
    \sum_{i=1}^{t}\log_2 \mathcal{L}_i(H')
    >
    \sum_{i=1}^{t}\log_2 \mathcal{L}_i(H).
\end{equation}
Each revision is checked against the trajectory collected so far within the current episode.
Thus, a new observation can suggest a revision, but accumulated evidence determines whether it replaces the current explanation. Role instance hypotheses that remain observationally indistinguishable are retained until subsequent interactions provide discriminating evidence.

\subsection{Role-Guided Decision Making}

Effective use of inferred role information involves both applying current role knowledge to decisions and checking uncertain assumptions through interaction. We use a reusable skill to guide the actor in both activities.

\paragraph{Organizing role information for the actor.}
The teacher examines the actor's interaction trajectories from the source world and develops a program that organizes role information around action requirements, predicted effects, and dependencies between actions. Together with predefined instructions for decision making and active verification, this program forms the skill.

At each step, the skill uses the current role instance hypotheses and their probabilities, observations, and interaction history to present relevant action candidates. The shared transition rules provide the requirements and predicted effects for these candidates under the current role instance hypotheses. Each interaction description draws its requirements and effects from the same combination of compatible role instance hypotheses. The skill also relates predicted effects to the prerequisites of subsequent actions and highlights uncertainty in role assignments and parameter estimates. The skill guides the actor to base judgments about object behavior on role instance hypotheses and interaction evidence, even when familiar names suggest different functions. The actor selects the final action from the environment's available actions, considering this information alongside its observations, task priorities, and immediate needs.

\paragraph{Active verification guided by the skill.}
A role instance hypothesis used for prediction may still lack interaction evidence, and support for its prototype assignment does not imply that all its parameter values or interactions have been verified. The skill encourages the actor to seek interaction evidence for unverified functions. Using the supplied role instance hypotheses, the actor assesses whether verification is useful for its decision and chooses which interaction, if any, to perform.

Feedback from the chosen action is used to refine role assignments and parameter estimates through the inference mechanism described above. The skill incorporates these updates and the interaction history into its guidance at the next step. In a target world, the teacher is absent, and both the actor's LLM parameters and the skill remain fixed.

\section{Experiment}

We evaluate \methodname{} in MARS \citep{tang2024mars}, an open-world survival benchmark built on Crafter \citep{hafner2021crafter}. Agents pursue 22 achievements in a partially observable world using text observations of nearby objects and their coordinates, the agent's status, and its inventory. MARS includes a \emph{Default} world and seven variants that alter terrain, survival settings, and task dependencies, individually or in combination, while retaining Crafter's objects and achievements. Changes such as hostile cows and stone yielding diamonds make object names unreliable indicators of behavior, testing adaptation to changed object functions. We compare \methodname{} with the MARS implementation of ReAct \citep{yao2023react,tang2024mars}, which reasons before each action; Induction from Reflection (IFR) \citep{tang2024mars}, which induces environment rules from trajectory reflections; and three world-model-based approaches: PoE-World \citep{piriyakulkij2026poe}, WALL-E \citep{zhou2026wall}, and WorldCoder \citep{tang2024worldcoder}.

\paragraph{Evaluation protocol.}
Table~\ref{tab:protocol} distinguishes three stages: \emph{Source learning} acquires knowledge in Default before transfer; \emph{Target learning} acquires knowledge in each target world before evaluation; and \emph{Online refinement} updates knowledge using feedback within each evaluation episode. We use two complementary comparisons to evaluate transfer. First, IFR, PoE-World, and WALL-E learn separately in each target world for five episodes before evaluation. This comparison tests whether \methodname{} can transfer knowledge from Default to achieve better task performance than baselines that learn in the target world. Second, WorldCoder and \methodname{} autonomously learn through interaction in Default, each with a source-learning budget of at most five episodes. Both enter target-world evaluation without target learning, providing a direct comparison of transfer performance. ReAct uses neither pre-evaluation learning stage.

All LLM calls during source learning use GPT-5.6 Sol \citep{openai2026gpt56}. WorldCoder constructs an executable world model. For \methodname{}, the teacher constructs role prototypes and transition rules, infers initial role assignments and parameter values, and develops the program that organizes role information for the actor. During target-world evaluation, WorldCoder refines its model code online, whereas \methodname{} refines only role assignments and environment-specific parameter estimates, keeping the prototypes, transition rules, and skill fixed. IFR continues to induce rules, PoE-World refines its experts and weights, and WALL-E refines symbolic knowledge.

We use Qwen3-8B \citep{yang2025qwen3} and GPT-5.6 Luna \citep{openai2026gpt56} as target-world backbones. In each configuration, the designated backbone handles all LLM calls during target learning and evaluation, including action selection and knowledge refinement. Episodes in all three stages are limited to 1,000 environment steps. Learned knowledge is retained across episodes within source learning and target learning. Each evaluation episode starts from the same knowledge state obtained before evaluation for that method and world; interaction history and refinements made during one evaluation episode are not carried into another.

\begin{table}[t]
    \centering
    \caption{Evaluation stages ($\checkmark$: yes; $\times$: no). Source learning: at most five episodes in Default; target learning: five per target world before evaluation; online refinement: within each evaluation episode.}
    \label{tab:protocol}
    \small
    \setlength{\tabcolsep}{5pt}
    \renewcommand{\arraystretch}{1.0}
    \begin{tabular}{@{}l*{6}{c}@{}}
        \toprule
        Stage & ReAct & IFR & PoE-World & WALL-E & WorldCoder & \methodname{} \\
        \midrule
        Source learning & $\times$ & $\times$ & $\times$ & $\times$ & $\checkmark$ & $\checkmark$ \\
        Target learning & $\times$ & $\checkmark$ & $\checkmark$ & $\checkmark$ & $\times$ & $\times$ \\
        Online refinement & $\times$ & $\checkmark$ & $\checkmark$ & $\checkmark$ & $\checkmark$ & $\checkmark$ \\
        \bottomrule
    \end{tabular}
\end{table}

We follow MARS \citep{tang2024mars} in computing episode reward and score (\%). For each target world, we evaluate with nine random seeds and report the mean and standard deviation. The \emph{All} entries report the arithmetic mean of the seven per-world means for each metric.

\begin{table}[t]
    \centering
    \caption{Reward and score (\%) on the seven MARS variants under the protocols in Table~\ref{tab:protocol}. Column groups indicate the backbone used for all target-world LLM calls.}
    \label{tab:main_results}
    \normalsize
    \setlength{\tabcolsep}{2pt}
    \renewcommand{\arraystretch}{1.05}
    \resizebox{\linewidth}{!}{%
    \begin{tabular}{@{}ll*{6}{c}*{6}{c}@{}}
        \toprule
        \multirow{2}{*}{Metrics} & \multirow{2}{*}{Mod. Type}
        & \multicolumn{6}{c}{Qwen3-8B}
        & \multicolumn{6}{c}{GPT-5.6 Luna} \\
        \cmidrule(lr){3-8}\cmidrule(lr){9-14}
        &
        & ReAct & IFR & \shortstack{PoE-\\World} & WALL-E
        & \shortstack{World\\Coder} & Ours
        & ReAct & IFR & \shortstack{PoE-\\World} & WALL-E
        & \shortstack{World\\Coder} & Ours \\
        \midrule
        \multirow{8}{*}{\textbf{Reward}}
        & Terrain     & $2.8 \pm 1.6$ & $2.0 \pm 0.9$ & $0.4 \pm 1.2$ & $2.0 \pm 2.2$ & $5.3 \pm 1.7$ & $\bm{9.4 \pm 3.7}$ & $4.0 \pm 2.5$ & $10.8 \pm 1.2$ & $0.8 \pm 0.7$ & $8.2 \pm 3.1$ & $5.8 \pm 1.7$ & $\bm{11.4 \pm 1.1}$ \\
        & Survival    & $2.0 \pm 1.2$ & $3.0 \pm 2.0$ & $0.0 \pm 1.1$ & $1.3 \pm 0.8$ & $2.3 \pm 1.4$ & $\bm{8.0 \pm 2.6}$ & $4.9 \pm 2.5$ & $9.3 \pm 1.5$ & $1.7 \pm 2.4$ & $10.2 \pm 4.0$ & $3.1 \pm 1.7$ & $\bm{10.8 \pm 2.4}$ \\
        & Task Dep.   & $1.9 \pm 1.7$ & $2.2 \pm 0.7$ & $1.5 \pm 1.6$ & $2.2 \pm 1.4$ & $0.8 \pm 0.8$ & $\bm{6.8 \pm 0.8}$ & $1.7 \pm 1.1$ & $7.7 \pm 1.0$ & $1.4 \pm 0.8$ & $4.8 \pm 2.5$ & $1.2 \pm 0.6$ & $\bm{8.5 \pm 1.5}$ \\
        & Terr. Surv. & $1.4 \pm 1.0$ & $1.7 \pm 0.8$ & $0.1 \pm 0.7$ & $1.7 \pm 3.2$ & $4.4 \pm 1.6$ & $\bm{8.1 \pm 3.3}$ & $4.1 \pm 1.5$ & $7.8 \pm 0.7$ & $2.1 \pm 0.9$ & $5.7 \pm 2.4$ & $4.3 \pm 1.4$ & $\bm{11.0 \pm 1.7}$ \\
        & Terr. Task. & $2.1 \pm 1.2$ & $3.0 \pm 0.7$ & $1.4 \pm 1.1$ & $1.7 \pm 1.4$ & $3.7 \pm 1.5$ & $\bm{6.2 \pm 2.1}$ & $3.8 \pm 1.9$ & $6.4 \pm 0.8$ & $3.8 \pm 0.8$ & $3.8 \pm 1.3$ & $4.2 \pm 1.4$ & $\bm{7.0 \pm 1.0}$ \\
        & Surv. Task. & $1.4 \pm 2.0$ & $3.2 \pm 1.5$ & $-0.6 \pm 0.5$ & $2.0 \pm 1.2$ & $1.3 \pm 0.6$ & $\bm{4.3 \pm 0.8}$ & $1.8 \pm 1.6$ & $3.3 \pm 1.6$ & $0.5 \pm 1.7$ & $4.3 \pm 2.2$ & $0.7 \pm 0.8$ & $\bm{4.7 \pm 0.8}$ \\
        & All Three   & $-0.2 \pm 0.6$ & $-0.1 \pm 0.6$ & $-0.5 \pm 0.5$ & $-0.2 \pm 0.8$ & $-0.0 \pm 0.6$ & $\bm{3.0 \pm 1.5}$ & $0.6 \pm 0.9$ & $2.5 \pm 1.5$ & $0.3 \pm 1.1$ & $0.9 \pm 1.5$ & $-0.4 \pm 0.5$ & $\bm{3.3 \pm 1.5}$ \\
        \cmidrule(l){2-14}
        & All         & 1.6 & 2.1 & 0.4 & 1.5 & 2.5 & \textbf{6.5} & 3.0 & 6.8 & 1.5 & 5.4 & 2.7 & \textbf{8.1} \\
        \midrule
        \multirow{8}{*}{\textbf{Score (\%)}}
        & Terrain     & $1.6 \pm 0.5$ & $1.5 \pm 1.1$ & $0.6 \pm 0.4$ & $1.7 \pm 1.2$ & $4.3 \pm 1.2$ & $\bm{16.6 \pm 9.4}$ & $3.3 \pm 1.0$ & $13.1 \pm 2.3$ & $0.6 \pm 0.1$ & $12.0 \pm 1.2$ & $5.5 \pm 0.5$ & $\bm{19.2 \pm 5.1}$ \\
        & Survival    & $1.5 \pm 0.2$ & $2.6 \pm 1.8$ & $0.5 \pm 0.4$ & $1.0 \pm 0.2$ & $1.5 \pm 0.3$ & $\bm{10.6 \pm 5.6}$ & $4.6 \pm 0.7$ & $14.7 \pm 3.5$ & $1.7 \pm 1.3$ & $16.3 \pm 0.8$ & $2.3 \pm 0.4$ & $\bm{19.1 \pm 7.4}$ \\
        & Task Dep.   & $1.5 \pm 0.6$ & $1.5 \pm 0.4$ & $1.2 \pm 0.6$ & $1.6 \pm 0.4$ & $0.6 \pm 0.2$ & $\bm{4.9 \pm 1.3}$ & $1.0 \pm 0.4$ & $7.1 \pm 2.0$ & $0.8 \pm 0.3$ & $4.0 \pm 1.3$ & $0.7 \pm 0.1$ & $\bm{9.9 \pm 4.8}$ \\
        & Terr. Surv. & $0.8 \pm 0.2$ & $1.1 \pm 0.6$ & $0.5 \pm 0.2$ & $1.0 \pm 0.2$ & $3.4 \pm 1.0$ & $\bm{11.0 \pm 4.8}$ & $2.3 \pm 0.5$ & $6.8 \pm 0.8$ & $1.4 \pm 0.3$ & $5.2 \pm 0.3$ & $2.8 \pm 1.1$ & $\bm{17.6 \pm 7.6}$ \\
        & Terr. Task. & $1.1 \pm 0.4$ & $1.8 \pm 0.6$ & $1.0 \pm 0.2$ & $1.2 \pm 0.3$ & $2.5 \pm 0.1$ & $\bm{5.8 \pm 2.9}$ & $2.6 \pm 1.0$ & $5.0 \pm 0.8$ & $2.3 \pm 0.4$ & $3.1 \pm 0.8$ & $2.7 \pm 0.3$ & $\bm{5.6 \pm 1.6}$ \\
        & Surv. Task. & $1.4 \pm 0.7$ & $2.8 \pm 1.7$ & $0.2 \pm 0.0$ & $1.5 \pm 0.3$ & $1.0 \pm 0.3$ & $\bm{4.4 \pm 1.7}$ & $1.5 \pm 0.4$ & $1.6 \pm 0.5$ & $0.9 \pm 0.7$ & $3.3 \pm 0.5$ & $0.6 \pm 0.1$ & $\bm{4.1 \pm 2.5}$ \\
        & All Three   & $0.3 \pm 0.1$ & $0.4 \pm 0.4$ & $0.2 \pm 0.0$ & $0.4 \pm 0.0$ & $0.3 \pm 0.1$ & $\bm{2.2 \pm 0.9}$ & $0.8 \pm 0.4$ & $1.6 \pm 1.0$ & $0.6 \pm 0.2$ & $1.1 \pm 0.5$ & $0.2 \pm 0.1$ & $\bm{2.4 \pm 0.5}$ \\
        \cmidrule(l){2-14}
        & All         & 1.2 & 1.7 & 0.6 & 1.2 & 1.9 & \textbf{7.9} & 2.3 & 7.1 & 1.2 & 6.4 & 2.1 & \textbf{11.1} \\
        \bottomrule
    \end{tabular}%
    }
\end{table}

As shown in Table~\ref{tab:main_results}, \methodname{} achieves the highest average reward across the seven variants with both backbones, demonstrating its effectiveness in supporting agent decision making across environments with altered dynamics. The advantage over WorldCoder under the same transfer protocol suggests that our role-based symbolic world model transfers more effectively across environments with changed object behavior.

\subsection{Ablation}

We evaluate four ablations with Qwen3-8B under the same transfer protocol: \emph{w/o Skill} retains online adaptation but removes the skill's information organization and decision/verification guidance; \emph{w/o Adaptation} freezes initial role assignments and parameter estimates; \emph{Fixed Role Assignments} permits parameter refinement but freezes object--role assignments; and \emph{w/o Active Verification} removes explicit verification guidance while retaining other decision support and online adaptation. Without the skill, the actor receives action requirements, predicted effects, and uncertainty from the highest-probability compatible combination of role instance hypotheses. We also compare \emph{Direct LLM}, which acts without a separate reasoning step or role guidance, and \emph{\methodname{} (Symbolic Planner)}, which replaces the actor with a hand-designed symbolic planner while retaining the same role representation and adaptation procedure. Implementation details appear in Appendices~\ref{sec:ablation_details} and~\ref{sec:decision_policy}.

\begin{table}[t]
    \centering
    \caption{Ablation results and decision-policy comparison on the seven MARS variants. LLM-based variants use Qwen3-8B. All is the arithmetic mean of the per-world means for each metric. Bold indicates the highest mean in each column within each metric.}
    \label{tab:ablation}
    \scriptsize
    \setlength{\tabcolsep}{2pt}
    \renewcommand{\arraystretch}{1.05}
    \resizebox{\linewidth}{!}{%
    \begin{tabular}{@{}ll*{7}{c}c@{}}
        \toprule
        Metrics & Method & Terrain & Survival & Task Dep. & Terr. Surv. & Terr. Task. & Surv. Task. & All Three & All \\
        \midrule
        \multirow{7}{*}{\textbf{Reward}}
        & Direct LLM         & $0.2 \pm 1.0$ & $2.0 \pm 1.7$ & $0.9 \pm 1.4$ & $1.3 \pm 1.2$ & $2.0 \pm 1.1$ & $0.8 \pm 1.4$ & $0.0 \pm 1.1$ & 1.0 \\
        & w/o Skill          & $0.9 \pm 0.8$ & $2.2 \pm 1.6$ & $1.3 \pm 1.0$ & $2.3 \pm 1.5$ & $1.1 \pm 0.9$ & $1.2 \pm 0.7$ & $-0.3 \pm 0.5$ & 1.2 \\
        & w/o Adaptation     & $9.7 \pm 1.8$ & $8.4 \pm 3.1$ & $0.9 \pm 0.4$ & $6.2 \pm 1.9$ & $4.3 \pm 2.4$ & $4.8 \pm 2.6$ & $0.2 \pm 1.4$ & 4.9 \\
        & Fixed Role Assignments & $6.8 \pm 3.1$ & $6.5 \pm 3.3$ & $1.4 \pm 1.3$ & $7.8 \pm 4.1$ & $5.4 \pm 1.4$ & $3.7 \pm 3.9$ & $2.5 \pm 1.7$ & 4.9 \\
        & w/o Active Verification & $8.0 \pm 1.9$ & $6.0 \pm 4.3$ & $1.2 \pm 0.7$ & $7.7 \pm 3.1$ & $1.7 \pm 1.5$ & $0.9 \pm 0.8$ & $-0.5 \pm 0.7$ & 3.6 \\
        & \methodname{} (full) & $9.4 \pm 3.7$ & $8.0 \pm 2.6$ & $6.8 \pm 0.8$ & $8.1 \pm 3.3$ & $6.2 \pm 2.1$ & $4.3 \pm 0.8$ & $3.0 \pm 1.5$ & 6.5 \\
        & \methodname{} (Symbolic Planner) & $\bm{13.5 \pm 0.6}$ & $\bm{14.9 \pm 1.5}$ & $\bm{8.3 \pm 1.8}$ & $\bm{12.4 \pm 1.6}$ & $\bm{14.2 \pm 1.7}$ & $\bm{15.8 \pm 1.9}$ & $\bm{6.5 \pm 2.6}$ & \textbf{12.2} \\
        \midrule
        \multirow{7}{*}{\textbf{Score (\%)}}
        & Direct LLM         & $0.5 \pm 0.3$ & $1.9 \pm 0.7$ & $1.0 \pm 0.4$ & $1.0 \pm 0.5$ & $1.5 \pm 0.0$ & $1.0 \pm 0.3$ & $0.5 \pm 0.4$ & 1.0 \\
        & w/o Skill          & $0.8 \pm 0.5$ & $1.4 \pm 0.6$ & $0.9 \pm 0.5$ & $2.1 \pm 1.7$ & $1.0 \pm 0.4$ & $1.0 \pm 0.2$ & $0.2 \pm 0.2$ & 1.0 \\
        & w/o Adaptation     & $12.6 \pm 3.8$ & $12.1 \pm 5.9$ & $0.8 \pm 0.3$ & $6.2 \pm 3.0$ & $2.8 \pm 2.1$ & $5.4 \pm 4.9$ & $0.6 \pm 0.6$ & 5.8 \\
        & Fixed Role Assignments & $7.7 \pm 1.7$ & $8.4 \pm 6.0$ & $1.2 \pm 0.3$ & $12.3 \pm 3.7$ & $3.6 \pm 0.4$ & $5.3 \pm 5.1$ & $1.8 \pm 0.8$ & 5.7 \\
        & w/o Active Verification & $9.8 \pm 3.8$ & $10.0 \pm 4.4$ & $0.9 \pm 0.2$ & $9.8 \pm 0.9$ & $1.3 \pm 0.3$ & $0.7 \pm 0.2$ & $0.2 \pm 0.1$ & 4.7 \\
        & \methodname{} (full) & $16.6 \pm 9.4$ & $10.6 \pm 5.6$ & $4.9 \pm 1.3$ & $11.0 \pm 4.8$ & $5.8 \pm 2.9$ & $4.4 \pm 1.7$ & $2.2 \pm 0.9$ & 7.9 \\
        & \methodname{} (Symbolic Planner) & $\bm{21.3 \pm 0.7}$ & $\bm{39.5 \pm 15.2}$ & $\bm{7.7 \pm 1.4}$ & $\bm{20.6 \pm 6.6}$ & $\bm{37.0 \pm 12.6}$ & $\bm{46.5 \pm 17.9}$ & $\bm{8.3 \pm 5.4}$ & \textbf{25.8} \\
        \bottomrule
    \end{tabular}%
    }
\end{table}

As shown in Table~\ref{tab:ablation}, the transferred role knowledge and skill provide a useful starting point for target-world decisions: \emph{w/o Adaptation} achieves an average reward of 4.9, compared with 1.0 for \emph{Direct LLM}. Online adaptation further increases average reward to 6.5, although freezing the source role estimates remains competitive in several variants. Thus, adaptation builds on a useful transferred prior, with benefits that vary across environments.

Although \emph{w/o Adaptation} and \emph{Fixed Role Assignments} both average 4.9 reward, per-world results suggest that parameter refinement and role reassignment are complementary, with neither consistently more important. On Terr. Surv., parameter refinement alone raises reward from 6.2 to 7.8, close to the full model's 8.1. On Task Dep., it raises reward only from 0.9 to 1.4, versus 6.8 with role reassignment. On Terr. Task., reward rises from 4.3 to 5.4 with parameter refinement and further to 6.2 with role reassignment. These patterns are consistent with mismatches in parameter values, functional bindings, or both. On All Three, parameter refinement accounts for most of the gain (0.2 to 2.5, versus 3.0 for the full model), suggesting that role reassignment's relative benefit does not necessarily increase as more environment factors change.

Realizing these benefits also requires guidance for using and testing role knowledge. Removing the skill reduces average reward to 1.2, close to \emph{Direct LLM}. Retaining its decision support without active verification raises reward to 3.6, but this remains below \emph{w/o Adaptation}. Together with the full model's 6.5, these results support the joint use of online adaptation and active verification. Verification provides larger reward gains in variants with task-dependency changes, consistent with uncertain assumptions about role instances and interaction requirements affecting downstream decisions through prerequisite chains.

The symbolic planner further illustrates the importance of how role knowledge is used. With the same role representation and adaptation procedure, it increases average reward from 6.5 to 12.2 and score from 7.9\% to 25.8\%. Together with the skill ablation, this suggests substantial room to improve the decision policy, whose actions also determine the feedback available for adaptation.

\subsection{Controlled Validation of Role Factorization}

We test whether interaction feedback can correct corrupted role estimates while keeping the environment, role prototypes, and transition rules fixed.

In \emph{Default}, we use a reference prior validated against environment action effects in controlled tests (Appendix~\ref{sec:reference_prior_validation}). It specifies 30 object--role assignments covering traversal, harvesting, crafting, and placement, with 110 role parameter fields. We corrupt the two separately: \emph{role parameter corruption} replaces 30\%, 50\%, 70\%, or 90\% of the fields with incorrect values drawn from their declared domains while retaining all role assignments; \emph{role assignment corruption} replaces the role types in the same proportions of the 30 assignments with randomly sampled incorrect types from the prototype library. We evaluate each condition after the first 100 and 300 active interactions from actor-generated trajectories, reporting means over six random seeds.

\emph{Role-type accuracy} is the percentage of the 30 reference object--role assignments whose inferred role types match the reference, irrespective of their parameter values. \emph{Role parameter accuracy} is the percentage of the 110 fields matching the reference values, counting missing fields and those associated with incorrect role assignments as incorrect. \emph{One-step prediction discrepancy} is the total variation distance between predicted and ground-truth distributions of direct action effects, averaged over 213 fixed, independent state--action pairs and reported as a percentage. The pairs cover satisfied and unsatisfied preconditions. Ground-truth distributions are obtained by executing the environment's action logic and enumerating its stochastic branches, excluding passive survival and NPC updates. If a required role assignment or prediction is unavailable, the discrepancy is set to 100\%. All conditions share this evaluation set, and evaluation does not update the model.

\begin{figure}[t]
    \centering
    \includegraphics[width=0.9\linewidth]{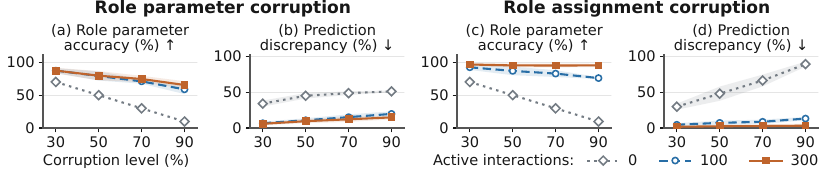}
    \caption{Controlled validation of role factorization. Role assignments and role parameter values are corrupted separately, while the environment, role prototypes, and transition rules remain fixed. Shaded bands indicate one standard deviation; see Appendix~\ref{sec:controlled_validation_details} for details.}
    \label{fig:robust}
\end{figure}

\begin{figure}[t]
    \centering
    \includegraphics[width=0.9\linewidth]{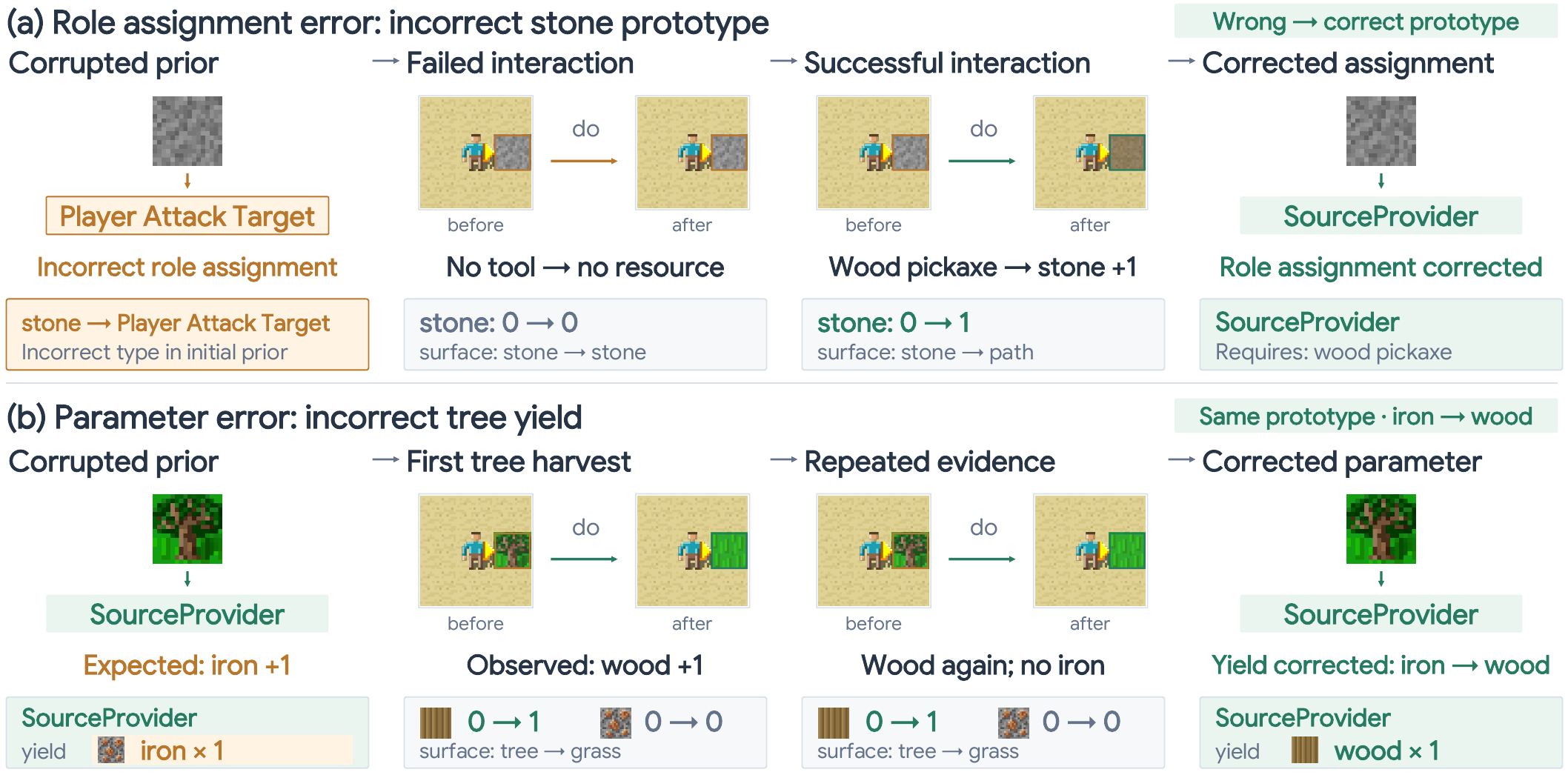}
    \caption{Correction of role estimates with fixed role prototypes and transition rules: (a) correcting a role assignment; (b) correcting a role parameter.}
    \label{fig:qualitative}
\end{figure}

Under role assignment corruption, role-type accuracy reaches 100\% by 100 active interactions and remains at 100\% at 300 interactions. Thus, role types are recovered within this interaction budget, while recovering the associated environment-specific parameters requires additional evidence. Figure~\ref{fig:robust} therefore focuses on role parameter accuracy and prediction discrepancy to characterize the remaining errors in role estimates and their predictive consequences.

Figure~\ref{fig:robust} shows greater recovery of role estimates under role assignment corruption than under parameter corruption at both interaction budgets and across all tested corruption levels. After 300 interactions, assignment corruption yields approximately 95--97\% role parameter accuracy and prediction discrepancy below 4\% across all corruption levels, suggesting limited sensitivity to initial role assignment errors within this controlled setting. A plausible explanation is that incorrect role types can produce qualitatively discriminative prediction mismatches, whereas identifying individual parameter values may require more specific interaction conditions.

Under 90\% parameter corruption, 300 interactions raise parameter accuracy from 10\% to approximately 65.5\%, while reducing prediction discrepancy from approximately 51\% to 15\%. Thus, exact recovery of role assignments and parameters is not a prerequisite for recovering useful predictions on the evaluated interactions. This suggests that adaptation can target role estimates that predict the relevant interaction dynamics without fully reconstructing the environment's symbolic model.

Figure~\ref{fig:qualitative} qualitatively illustrates role estimate correction. For stone in Figure~\ref{fig:qualitative}(a), failure without a tool followed by successful harvesting with a wooden pickaxe supports replacing the illustrative \emph{Player Attack Target} assignment with \emph{SourceProvider}. For tree in (b), repeated observations of wood rather than iron correct the resource-yield parameter while preserving its \emph{SourceProvider} assignment.

\section{Conclusion}

We introduced \methodname{}, which transfers world knowledge through functional role abstractions, reusing role prototypes and transition rules while refining environment-specific role assignments and parameter estimates through interaction feedback. Across both backbones in MARS, transferring from Default yields higher average reward than the evaluated baselines under the reported configurations. Controlled interventions further show that useful predictions can be recovered by correcting role estimates while keeping shared prototypes and rules fixed.

\section*{Acknowledgement}
This work was supported by National Natural Science Foundation of China (U23A20311).

\bibliographystyle{named}
\bibliography{reference}

\appendix
\clearpage
\section{Appendix}
\subsection{Offline World Model Construction Details}
\label{sec:offline_construction_details}

\paragraph{Inputs and predefined representation.}
The teacher constructs role prototypes and transition rules from source-world trajectories and additional interaction evidence. The role representation, DSL event vocabulary, condition and effect operators, and interpreter are provided in advance. This includes the execution semantics and event scheduling used to compose rules. Within this representation, the teacher selects the functional decomposition, declares parameters and state variables, and composes conditions and effects. These structural choices are made by the teacher; the interpreter executes the resulting rules. Environment-interaction tools allow the teacher to restore states, set controlled conditions, and execute actions. Evidence for induction consists of public observations before and after each action, together with the recorded intervention conditions.

\paragraph{Preparing interaction evidence.}
The teacher compares observations before and after each action to identify candidate effects, retaining unsuccessful attempts and interactions with no observable change. For controlled interactions, the observation before the action is recorded after setting the experimental conditions, so changes made by the tools are excluded from the action's effects. Comparisons with control conditions and no-operation actions help distinguish action effects from background changes. An observed difference alone does not establish which mechanism caused it. Likewise, an unchanged observation may reflect unmet conditions or a stochastic outcome, so it does not by itself eliminate a candidate function.

\paragraph{Jointly proposing prototypes and transition rules.}
For each candidate mechanism, the teacher selects a triggering event $\mathrm{event}$ from the predefined vocabulary and specifies participant variables and their required role prototypes in $P_{\tau}$. Successful and unsuccessful interactions constrain the applicability condition $g_{\tau}$. The effects $F_{\tau}$ specify the resulting state changes and interaction outcomes, including dependencies among effects and conditions under which they occur. Effects that share a stochastic outcome are represented jointly, preserving which changes occur together. Outcome probabilities are estimated from repeated interactions under the corresponding conditions; observations of only successful outcomes support a deterministic modeling assumption, rather than proving that the underlying probability is one.

The teacher synchronizes these rules with the associated prototypes. Parameters describe interaction requirements and effects that are fixed for an object within an environment, with values in $\Theta_{\rho}$; state variables describe properties that evolve through interaction, with values in $\mathcal{X}_{\rho}$. Candidate parameter values are supported by the interaction evidence. Each declared parameter or state variable must be referenced by at least one transition condition or effect. Thus, constructing a prototype includes determining the variables required by its rules, and revising a rule may require corresponding changes to the prototype.

\paragraph{Choosing the scope of a revision.}
The teacher first checks whether a discrepancy can be explained by parameter values within the current prototype and transition structure. Assigning an existing prototype to another object also requires no new functional definition. If the same function instead needs an additional applicability condition or a revised outcome structure, the teacher modifies its transition rules and any required parameter or state declarations. A new or split prototype is introduced when an independent function requires a separate description of its events, participants, conditions, or effects; the associated transition rules are constructed at the same time. A failed parameter revision therefore does not automatically imply a new prototype. Functions that coexist independently on the same object remain separate, with $\mathcal{C}_{\rho}$ specifying which prototypes may be instantiated together.

\paragraph{Parameterizing and merging candidates.}
The teacher merges candidates when their triggering events, participant structure, and condition--effect structure can be shared, including the temporal relationships and dependencies among outcomes. Object-specific constants become parameters, supported values enter the parameter domains, and each object's complete parameter combination is retained in its source-world role instance. The associated transition rules are merged at the same time: prototype references, participant requirements, and parameter or state references are updated, and superseded rules are removed to avoid executing duplicate effects. Similar outputs or coexistence on the same object are insufficient grounds for a merge. Distinct conditions or temporal relationships must remain explicit as conditional branches or separate functions.

\paragraph{Executing rules and checking revisions.}
The interpreter executes candidate rules using the observed state before the action and the parameter values of the corresponding role instances, composing applicable rules according to the predefined DSL semantics. Predictions are compared with the complete observable effects covered by the candidate model, including quantities or fields expected to remain unchanged. Deterministic predictions are checked for agreement with these effects. For stochastic predictions, checking whether an observation lies in the predicted support is distinct from comparing predicted probabilities with empirical outcome frequencies. Missing effects, extra effects, incorrect conditions, and incorrect dependencies among outcomes provide counterexamples for revising the model.

After a structural revision, the teacher checks the affected rules against accumulated successful and unsuccessful interactions. Parameterization and merging are also checked for preservation of the predicted outcome distributions on the recorded inputs. This check evaluates whether the revised representation preserves the candidate mechanisms on those inputs; it supplies no independent evidence for their correctness on unobserved inputs. Additional controlled interactions can distinguish competing explanations. The model is frozen before checking it on held-out interactions, which are not used to select its structure or estimate its parameters.

\paragraph{Retained model and transfer.}
The retained prototypes define reusable functions and their parameter and state spaces; the transition rules give their executable behavior. Source-world role assignments and parameter values are retained separately as the initial role prior, using the formats in Appendix~\ref{sec:representation_formats}. In the target world, prototypes and transition rules remain fixed while role assignments and parameter estimates are refined online. The construction establishes an evidence-supported representation within the observed conditions and the supplied DSL; finite trajectories need not uniquely determine the functional decomposition.

\subsection{Role Prototype, Transition, and Prior Formats}
\label{sec:representation_formats}

We provide example formats for role prototypes, transition rules, and initial role priors below. The schemas below show their field names and nesting, with angle-bracketed placeholders describing the corresponding contents. Each array may contain multiple entries of the displayed form.

\paragraph{Role prototypes.}
Each role prototype defines a reusable function, the compatible prototypes that may be instantiated on the same object, and the parameter and state spaces of its instances. Parameters describe properties that are fixed for a given object within an environment; state variables track quantities that can change during interaction.
\begin{shaded*}
\small
\begin{verbatim}
{
  "role_prototypes": [
    {
      "prototype": {
        "id": "<Unique prototype identifier>",
        "name": "<Human-readable prototype name>",
        "definition": "<Reusable behavioral function>",
        "compatible_prototype": ["<Compatible prototype ID>"],
        "decision_invariants": ["<Criteria for role inference>"],
        "variables": {
          "parameters": {
            "<parameter_name>": {
              "type": "<Parameter data type>",
              "domain": <Allowed values or numeric interval>
            }
          },
          "state": {
            "<state_name>": {
              "type": "<State variable data type>"
            }
          }
        }
      }
    }
  ]
}
\end{verbatim}
\end{shaded*}
A parameter's \texttt{domain} is specified by an explicit list of admissible \texttt{values} or an \texttt{interval} of numeric bounds. The \texttt{parameters} and \texttt{state} mappings may be empty when the prototype declares no variables of that kind.
\par\medskip

\paragraph{Transition rules.}
Each rule specifies a triggering event, participant aliases with their required role prototypes, an applicability condition, and effects. Participant aliases bind concrete role instances when evaluating the DSL condition and effects.
\begin{shaded*}
\small
\begin{verbatim}
{
  "transitions": [
    {
      "id": "<Unique transition identifier>",
      "role": "<Prototype ID associated with this rule>",
      "event": "<Event triggering rule evaluation>",
      "participants": {
        "<participant_alias>": "<Required prototype ID>"
      },
      "when": <DSL condition for applying the rule>,
      "effects": [<DSL operations for state changes and outcomes>]
    }
  ]
}
\end{verbatim}
\end{shaded*}
Each effect operation specifies \texttt{op} (the operation), \texttt{observable} (whether it contributes observable evidence), and \texttt{evidence\_owner} (the participant to which the evidence is attributed), together with operation-specific arguments.
\par\medskip

\paragraph{Initial role priors.}
The prior groups initial object--role assignments by prototype identifier and supplies environment-specific parameter values for each binding. These values instantiate the parameter declarations in the corresponding prototype.
\begin{shaded*}
\small
\begin{verbatim}
{
  "role_bindings": [
    {
      "role_id": "<Assigned prototype identifier>",
      "instance_bindings": [
        {
          "observed_type": "<Object type assigned this prototype>",
          "parameters": {
            "<parameter_name>": <Initial parameter value>
          }
        }
      ]
    }
  ]
}
\end{verbatim}
\end{shaded*}
The prototypes and transition rules remain fixed in the target world. The prior initializes role assignments and parameter estimates, which are revised online from interaction feedback; current state values are supplied separately during interaction.

\subsection{Ablation Details}
\label{sec:ablation_details}

\paragraph{Shared settings.}
The four component ablations in Table~\ref{tab:ablation} use Qwen3-8B in the seven non-Default MARS worlds, with the same source-world prior, role prototypes, and executable transition rules as the full method. Each episode resets role beliefs and interaction history. The prototypes, transition rules, and LLM parameters remain fixed throughout evaluation. All variants use the same observation format, up to ten recent action--observation turns, and the same action set and decoding settings. Episodes are limited to 1,000 environment steps, including an initial \texttt{noop}; an episode reaching this limit therefore contains at most 999 actor decisions.

\paragraph{w/o Skill.}
We retain online role inference and refinement and query the same transition model for available actions, including hypothetical \texttt{do} interactions after approaching and facing other visible object types. For each query, a fixed filter selects the compatible combination of role instance hypotheses with the highest probability mass. It presents preconditions, effect quantities and conditional probabilities, current outcome predictions, verification status, and unknowns. Hypothesis probability mass and action-effect probabilities are displayed separately. Rules with mismatched action conditions or no effects are omitted, together with internal parameter tables and repeated evidence. We remove both the program-generated decision guidance and the fixed textual skill instructions. The remaining presentation does not rank information by task or history, resolve prerequisite chains, or provide navigation and retry suggestions. The actor still receives its observation and action-feedback history, and its chosen actions continue to supply evidence for online adaptation.

\paragraph{w/o Adaptation.}
At every decision, we construct the role-query interface from a fresh copy of the initial source-world prior and populate it with the current observed state. Interaction evidence does not revise role assignments, parameter hypotheses, their weights, or verification status. Prior assignments retain their initial unverified status. The full skill, including its active-verification guidance, remains available. Current observations, inventory, status, observed movement, and action history are still supplied, allowing the skill to respond to changed conditions and previous failures. Thus, the frozen component is the persistent role knowledge; the decision input continues to reflect the ongoing episode.

\paragraph{Fixed Role Assignments.}
We fix each object's assigned prototype set to its initial assignments and disable the addition, removal, and replacement of assigned prototypes. Interaction feedback still refines parameter values and hypothesis weights within the assigned prototypes. For example, the model can revise the yield or requirements of an assigned resource-provider role instance. In this implementation, role verification status and event-support flags also remain at their initial values. The full skill, current observations, action history, and verification-related information are retained. This condition therefore permits parameter refinement within the prior assignments while keeping the functional assignments and their verification status fixed.

\paragraph{w/o Active Verification.}
We retain the skill's ordinary decision support and full online adaptation but remove its explicit support for active verification. In the full method, the skill provides role instance hypotheses and their probabilities, verification status, interaction evidence, and action requirements and predicted effects. Using this information alongside its observations and task priorities, the actor decides whether verification is useful and which interaction, if any, to perform. Ordinary action descriptions in the ablation use the highest-probability compatible candidate. Prerequisite planning, exploration to find missing materials or stations, risk indications, and failure history are retained. The actor remains free to choose any available action, including an uncertain interaction, and feedback from its actions continues to update role assignments and parameter estimates. This ablation evaluates the contribution of the skill's support for the actor's own verification decisions.

\subsection{Symbolic Planner Settings}
\label{sec:decision_policy}

For \methodname{} (Symbolic Planner), we replace the LLM actor with a hand-designed symbolic planner that operates on inferred role instances. The role prototypes, executable transition rules, and online role inference and refinement procedure are retained. We evaluate this variant on the same seven MARS variants, with reward and score reported in Table~\ref{tab:ablation}.

\paragraph{Role-based action representation.}
At each step, the planner associates observed objects and remembered terrain with their current role assignments. For ordinary planning, each assigned prototype is instantiated with its highest-probability parameter hypothesis. The shared transition rules then provide action descriptions consisting of an action, its preconditions, and its predicted effects. These include resource requirements, proximity to workstations, inventory and status changes, object creation, and achievement progress. The planner interprets the DSL condition and effect vocabulary; prototype identifiers select the corresponding inferred role instances for each object. Thus, concrete object functions and interaction requirements enter planning through the current role estimates.

\paragraph{Goal selection and prerequisite planning.}
A fixed goal scheduler prioritizes survival and recovery, followed by unfinished achievements. The planner matches a desired outcome to a role-instantiated transition effect and recursively resolves unmet prerequisites. For example, to obtain a resource, it finds an interaction predicted to increase that resource, then plans how to acquire any required tools or establish access to a required workstation. Each dependency is read from the transition conditions and linked to another action through its predicted effects. Resource and achievement plans are compared using action cost and the predicted probability of the desired effect, while a feasible ongoing achievement target is retained. Navigation searches over position and facing on a map accumulated from observations, using role-derived traversability and risk estimates from inferred role instances and interaction history.

\paragraph{Active verification and closed-loop execution.}
When ordinary goal plans are blocked, the planner can test goal-relevant role instance hypotheses or unverified interaction prerequisites. It compares candidate predictions of observable outcomes and ranks informative tests by expected information gain per action cost, subject to survival constraints. When no goal or verification plan is available, it explores the frontier of the observed map. The planner executes only the first action of each plan, records its prediction before execution, and compares it with the resulting feedback. The same interaction feedback updates role assignments and parameter estimates through the online inference procedure. The next decision uses the refreshed role information to compute a new plan. Throughout this process, the role prototypes and transition rules remain fixed, while revised role estimates can change both the selected objects and the prerequisite chains used to reach a goal.

\subsection{Controlled Validation Details}
\label{sec:controlled_validation_details}

\paragraph{Reference and corruption settings.}
The experiment covers 30 object--role assignments in Default: 13 for traversal, seven for harvesting, six for crafting, and four for placement. Together they contain 110 parameter fields. The reference prior is validated in Appendix~\ref{sec:reference_prior_validation}; the environment, role prototypes, and transition rules remain fixed during adaptation.

For parameter corruption, the four levels alter 33, 55, 77, or 99 fields while retaining all bindings. Fields are sampled without replacement from the 104 fields that admit an incorrect value in their declared domains, including action-valued fields; the six fields with singleton domains remain unchanged. Each selected field receives a uniformly sampled incorrect value. The role assignment corruption protocol selects 9, 15, 21, or 27 assignments. In each selected assignment, the correct role type is replaced by a randomly sampled incorrect type from the prototype library, and the replacement prototype's parameters are sampled independently from their respective declared domains. Unselected assignments retain their reference role types and parameter values. For each of six seeds, a random ordering of eligible fields or assignments defines nested corruption sets, so a higher level includes the changes at lower levels. Each condition starts from its own corrupted copy of the reference prior.

\paragraph{Active-interaction budget and evidence.}
We count active interactions by the agent's verification attempts. Each \emph{active interaction} consists of an action selected by the agent to test an uncertain role assignment, parameter value, or interaction prerequisite, together with its observed outcome. Actions used only for navigation or routine resource acquisition do not contribute to this count. A movement action selected to verify traversability is counted, and both successful and unsuccessful verification attempts are retained.

We let the agent continue interacting until the required number of active verification attempts has been reached, and extract the observation--action--observation records from the first 100 or 300 attempts in chronological order. These records are processed sequentially through the same online inference and refinement procedure to evaluate how role assignments and parameter estimates change as verification feedback accumulates. The role prototypes and transition rules remain fixed.

\paragraph{Recovery metrics.}
Role parameter accuracy uses all 110 reference fields as a fixed denominator, with equal weight per field. Missing fields and all reference fields associated with a missing or incorrect role assignment count as incorrect. A list-valued parameter is evaluated as one field; order is ignored for resource requirements, resource costs, workstation requirements, and allowed surfaces. When several parameter hypotheses share the highest probability, a field counts as correct only if it matches the reference in every tied hypothesis. Role-type accuracy is evaluated over all 30 reference object--role assignments and checks whether the inferred role types match the reference, independently of parameter accuracy.

\paragraph{Independent prediction evaluation.}
All frozen models are evaluated on the same set $\mathcal{Q}$ of 213 state--action pairs: 38 traversal, 57 harvesting, 42 crafting, and 76 placement cases. These are constructed separately from the adaptation corpus and cover tool combinations, material thresholds, workstation availability, surface types, and occupancy constraints. Each case evaluates its corresponding binding using the highest-probability parameter hypothesis; ties select the first stored candidate. Auxiliary quantity constraints are held at their reference values. The compared effects comprise inventory and status increments, achievement increments, player displacement, and changes to the action's target cell. The environment distribution is obtained by executing its action logic and enumerating stochastic branches, excluding passive survival and NPC updates while retaining quantity limits and lethal-entry effects.

For a test pair $q$, let $P_t(y\mid q)$ denote the frozen model's distribution over direct effects after $t$ active interactions, and let $P_{\mathrm{env}}(y\mid q)$ denote the environment distribution. We report
\begin{equation}
    E_{\mathrm{pred}}(t)
    = \frac{100}{213}\sum_{q\in\mathcal{Q}}
      \frac{1}{2}\sum_y
      \left|P_t(y\mid q)-P_{\mathrm{env}}(y\mid q)\right|.
\end{equation}
An unavailable binding or incomplete prediction receives a per-case error of one and remains in the denominator. All cases receive equal weight, and evaluation never updates the model. This metric measures direct effects of individual bindings under the specified test conditions.

\paragraph{Aggregation and reported checkpoints.}
Table~\ref{tab:controlled_validation_details} reports means and sample standard deviations across the six seeds, with standard deviations computed using a denominator of five. In Figure~\ref{fig:robust}, shaded bands show the mean plus or minus one standard deviation. Prediction discrepancy is evaluated at 0, 100, and 300 active interactions. The plotted initial parameter accuracy is $100\%$ minus the corruption percentage: this is exact for parameter corruption and a nominal reference for role assignment corruption, since reference assignments contain different numbers of fields. All post-adaptation accuracies use the actual field counts.

\begin{table}[t]
    \centering
    \caption{Controlled validation results (mean $\pm$ sample standard deviation across six random seeds). Column subheadings give the active-interaction budget; both metrics are reported as percentages.}
    \label{tab:controlled_validation_details}
    \small
    \setlength{\tabcolsep}{3pt}
    \renewcommand{\arraystretch}{1.15}
    \begin{tabular}{@{}llrrrrr@{}}
        \toprule
        Corruption type & Level (\%) & \multicolumn{2}{c}{Role parameter accuracy} & \multicolumn{3}{c}{Prediction discrepancy} \\
        \cmidrule(lr){3-4}\cmidrule(lr){5-7}
        & & 100 & 300 & 0 & 100 & 300 \\
        \midrule
        \multirow{4}{*}{Role parameters}
        & 30 & $87.27 \pm 3.77$ & $87.27 \pm 4.23$ & $34.19 \pm 4.22$ & $6.81 \pm 2.09$ & $6.00 \pm 1.93$ \\
        & 50 & $79.09 \pm 6.88$ & $79.55 \pm 7.01$ & $45.15 \pm 3.84$ & $10.95 \pm 2.83$ & $9.68 \pm 2.80$ \\
        & 70 & $70.91 \pm 3.50$ & $74.55 \pm 4.78$ & $48.79 \pm 2.59$ & $15.33 \pm 3.84$ & $12.30 \pm 3.56$ \\
        & 90 & $58.79 \pm 7.10$ & $65.45 \pm 5.75$ & $51.29 \pm 1.25$ & $19.85 \pm 3.53$ & $14.98 \pm 2.81$ \\
        \midrule
        \multirow{4}{*}{Role assignments}
        & 30 & $92.27 \pm 3.93$ & $96.52 \pm 2.11$ & $29.89 \pm 4.26$ & $4.69 \pm 2.18$ & $1.88 \pm 1.92$ \\
        & 50 & $86.82 \pm 6.50$ & $95.15 \pm 2.74$ & $48.04 \pm 9.91$ & $7.36 \pm 3.12$ & $2.59 \pm 2.02$ \\
        & 70 & $82.73 \pm 5.17$ & $95.00 \pm 2.80$ & $66.43 \pm 8.21$ & $8.84 \pm 3.00$ & $2.85 \pm 2.53$ \\
        & 90 & $75.91 \pm 1.38$ & $95.15 \pm 2.48$ & $89.28 \pm 3.48$ & $13.34 \pm 2.27$ & $3.39 \pm 2.69$ \\
        \bottomrule
    \end{tabular}
\end{table}

\subsection{Environment-Transition Validation of the Reference Prior}
\label{sec:reference_prior_validation}

\paragraph{Scope and prediction targets.}
We independently evaluate the uncorrupted reference prior used to construct the controlled corruption experiments. The evaluation covers 30 object--role assignments in Default: 13 traversal assignments (R0), seven harvesting assignments (R1), six crafting assignments (R2), and four placement assignments (R3), together with the auxiliary quantity constraints (R12) used in their predictions. Resources and tools have a maximum quantity of nine. The R0--R3 reference parameters are constructed from the environment configuration, so agreement with those fields alone is not independent evidence of predictive performance. We instead compare predictions with effects obtained by executing the environment. The prior and transition rules remain frozen throughout evaluation, with no learning or LLM-based revision.

For each state--action pair, the predictor uses only information visible before the action. We compare the complete vector of inventory and vital-statistic increments, achievement-count increments, player displacement, changes to the action's target cell, and the player's death status, including fields expected to remain unchanged. Occupation of the target cell by the player after movement is not counted as a terrain change. Changes to other cells, full NPC states, rewards, and hidden timers are outside the prediction target.

\paragraph{Controlled scenarios.}
We evaluate 585 state--action pairs, retaining all 213 inputs from the main corruption experiment and adding condition boundaries and combinations. Test scenarios are generated from environment definitions without reading the prior under evaluation. They cover traversal directions, health levels, occupied cells and boundaries; harvesting with different tool combinations and inventory quantities; crafting with insufficient, exact, or surplus materials and different workstation combinations and distances; placement across terrain types, material quantities, and occupancy conditions; and repeated triggering of previously unlocked achievements.

Each case restores a complete snapshot from normal environment initialization, then sets the required terrain, inventory, workstations, and occupancy. NPCs are removed; occupancy tests use immature plants. Vital-statistic timers retain their initial values and the environment remains at step zero. We execute a full environment step without replacing action, background-update, or random-number functions. Background updates still run, but under these initial conditions they do not change the compared vital statistics during that step. Thus, these are controlled interventions rather than samples from the natural state distribution. Deterministic cases run once. Each stochastic case runs 1,000 times from the same physical state with different environment random seeds. The evaluation comprises 566 deterministic cases and 19 stochastic cases, requiring 19,566 environment steps. The deterministic cases comprise 155 traversal, 115 harvesting, 72 crafting, and 224 placement cases.

\paragraph{Metrics.}
For deterministic cases, a match requires agreement on the entire visible effect. For a state--action pair $q$, stochastic predictions are compared with empirical outcome frequencies using total variation (TV):
\begin{equation}
    \widehat{E}(q)=\frac{1}{2}\sum_y\left|P_{\mathrm{ref}}(y\mid q)-\widehat{P}_{\mathrm{env}}(y\mid q)\right|.
\end{equation}
We compute two-sided 95\% Wilson intervals for each outcome frequency. These are pointwise intervals, not a joint acceptance criterion; finite-sample empirical TV need not be zero even for a matching distribution. We report both the mean over equally weighted state--action pairs and the mean obtained by averaging within each assignment and then equally across the 30 assignments. Repeated stochastic trials improve probability estimates without increasing a case's aggregation weight. Cases without a prediction remain in the denominator and receive TV of one.

\paragraph{Results and sensitivity to injected errors.}
The reference prior produces complete predictions for all 585 controlled cases and matches all 566 deterministic cases, including harvesting and crafting at inventory capacity. Its case-averaged and assignment-averaged empirical TV are 0.0410\% and 0.0421\%, respectively (Table~\ref{tab:reference_prior_validation}).

All 19 stochastic cases involve grass harvesting with a predicted drop probability of 50\%. A drop event is identified by an increment in the \texttt{collect\_sapling} achievement count; at full inventory, the event can occur without increasing the sapling quantity. Across 19,000 trials, 9,602 drop events occur, giving a pooled frequency of 50.5368\% with a 95\% Wilson interval of $[49.8259\%, 51.2476\%]$. Per-case frequencies range from 47.8\% to 52.9\%, and the predicted probability lies within the corresponding pointwise 95\% interval in all 19 cases. Mean empirical TV over these cases is 1.2632\%. These results support using the prior as an environment-consistent reference within the tested states, conditions, and visible effects; they do not establish correctness for other role assignments, all reachable states, or complete world-state predictions.

To check whether the evaluation can reveal errors, we independently introduce seven single changes to the reference prior: remove the tree/R1 assignment; require a wooden pickaxe for harvesting trees; change the wood cost of a wooden pickaxe to two; remove the workstation requirement for crafting an iron pickaxe; allow planting on sand; change the grass drop probability from 0.5 to 0.1; or change the drink quantity limit to eight. Every model is evaluated against the same environment outcomes. The deterministic changes produce counterexamples or missing predictions. The drop-probability change raises mean TV over the 19 stochastic cases from 1.2632\% to 40.5368\%. All seven error types are detected through their corresponding scenarios.

\begin{table}[!htbp]
    \centering
    \caption{Environment-transition validation of the frozen reference prior and independent single-error controls. Deterministic mismatches are counted over 566 cases and include missing predictions. Both TV averages cover all 585 cases and are reported as percentages.}
    \label{tab:reference_prior_validation}
    \small
    \setlength{\tabcolsep}{5pt}
    \renewcommand{\arraystretch}{1.15}
    \begin{tabular}{@{}lrrrr@{}}
        \toprule
        Prior & \shortstack{Deterministic\\mismatches} & \shortstack{Missing\\predictions} & \shortstack{Case\\mean TV} & \shortstack{Assignment\\mean TV} \\
        \midrule
        Reference & 0 & 0 & 0.0410 & 0.0421 \\
        Remove tree/R1 & 19 & 19 & 3.2889 & 3.3754 \\
        Require tool for tree & 12 & 0 & 2.0923 & 2.1474 \\
        Change wood cost & 7 & 0 & 1.2376 & 2.1633 \\
        Remove workstation requirement & 4 & 0 & 0.7248 & 1.0677 \\
        Allow planting on sand & 4 & 0 & 0.7248 & 0.2802 \\
        Change drop probability & 0 & 0 & 1.3166 & 1.3512 \\
        Change drink quantity limit & 3 & 0 & 0.5538 & 0.5421 \\
        \bottomrule
    \end{tabular}
\end{table}

\paragraph{Natural-trajectory diagnostic.}
We additionally collect six trajectories from normal spawn states using independent environment seeds and uniformly random actions, with a limit of 300 steps per trajectory and termination on death. Resources are not prepared, NPCs remain present, and the prior neither selects actions nor learns from feedback. The trajectories contain 1,140 transitions and visit 15 of the 30 assignments: R0--R3 predictions are available for 787 transitions, and 113 of these have observed effects outside the predicted support. The remaining 353 transitions comprise 222 steps in sleeping states, 122 actions outside the 30 assignments, and nine object interactions. We retain all transitions, including background changes and cases without predictions. The mean multiclass Brier score is 0.825439 when unavailable predictions receive the maximum penalty of two. This diagnostic reflects both prediction quality and coverage of R0--R3; its discrepancies cannot be interpreted directly as parameter errors in these assignments or as a measure of the entire prior's predictive accuracy, since joint prediction with other role instances is not implemented here.

\end{document}